\documentclass[runningheads]{llncs}
\usepackage[T1]{fontenc}
\usepackage{graphicx}

\usepackage{amsmath,amssymb} 
\usepackage{color}
\usepackage{hyperref}
\usepackage{makecell}
\usepackage{subcaption}
\begin{document}
\title{Neural Residual Flow Fields\\ for Efficient Video Representations}
\titlerunning{NRFF}

\author{Daniel Rho\inst{1}\orcidID{0000-0002-8568-9489} \and
Junwoo Cho\inst{1} \and
Jong Hwan Ko\inst{1,2}\thanks{Corresponding authors.}\and
Eunbyung Park\inst{1,2}$^{\star}$}

\authorrunning{D. Rho et al.}
\institute{Department of Artificial Intelligence, Sungkyunkwan University\\ \and
Department of Electrical and Computer Engineering, Sungkyunkwan University
\email{\{daniel231,jwcho000,jhko,epark\}@skku.edu}}

\maketitle

\begin{abstract}
Neural fields have emerged as a powerful paradigm for representing various signals, including videos.
However, research on improving the parameter efficiency of neural fields is still in its early stages.
Even though neural fields that map coordinates to colors can be used to encode video signals, this scheme does not exploit the spatial and temporal redundancy of video signals.
Inspired by standard video compression algorithms, we propose a neural field architecture for representing and compressing videos that deliberately removes data redundancy through the use of motion information across video frames. 
Maintaining motion information, which is typically smoother and less complex than color signals, requires a far fewer number of parameters.
Furthermore, reusing color values through motion information further improves the network parameter efficiency.
 In addition, we suggest using more than one reference frame for video frame reconstruction and separate networks, one for optical flows and the other for residuals.
Experimental results have shown that the proposed method outperforms the baseline methods by a significant margin.
The code is available in \url{https://github.com/daniel03c1/eff_video_representation}.
\end{abstract}


\begin{figure*}[t]
\centering
\includegraphics[width=1\textwidth]{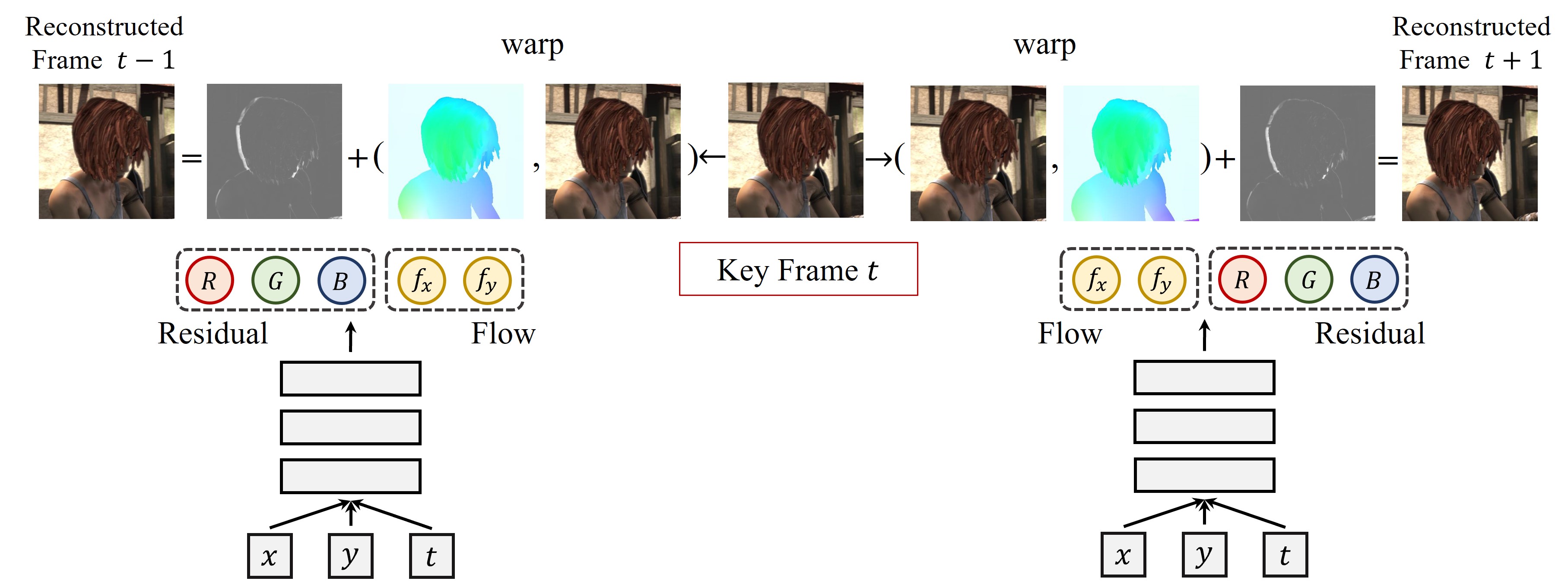}
 \caption{Neural residual flow fields (NRFF) with a single reference frame.
The gray-scale image on each side shows residuals, and the blue image in each parenthesis are optical flows. x, y, t are spatial and temporal coordinates.}
\label{fig:main}
\end{figure*}

\section{Introduction}

Neural fields~\cite{tancik2020fourier,sitzmann2020implicit,meta-implicit,martel2021acorn,mildenhall2020nerf,streamable} (also known as implicit neural representations or coordinate-based neural representations) are an emerging approach for representing various signals.
Signals can be reconstructed using dense sampling once a neural network has been trained to map coordinates to corresponding signal values.
Unlike other representation techniques that store discretely sampled data, neural fields use continuous coordinates as inputs, allowing them to represent signals at any resolution and at any arbitrary coordinate.
It can accurately express both low and high frequencies of signals thanks to recent breakthroughs in input features~\cite{tancik2020fourier,sitzmann2020implicit,mildenhall2020nerf}.
This innovative representation approach has shown considerable promise in a number of areas, including computer graphics~\cite{mildenhall2020nerf,mescheder2019occupancy}, physical simulations~\cite{sitzmann2020implicit,raissi2019physics}, and generative models~\cite{adversarial-implicit,generative-implicit}, to name a few.

Although it has recently received widespread attention, its parameter efficiency has not been thoroughly investigated.
In recent studies, neural fields require a large number of parameters to accurately represent signals~\cite{sitzmann2020implicit,martel2021acorn,mildenhall2020nerf}.
Without enhancing parameter efficiency, transaction costs of neural fields are high because signals are stored as network parameters.
This hinders us from utilizing it in many practical applications that may benefit from it.

In this paper, we study how to effectively represent videos using this new representation approach.
A naive approach, equivalent to SIREN~\cite{sitzmann2020implicit}, would be to use a neural field as a function of spatial and temporal coordinates, $(r, g, b) = f_\theta(t, x, y)$, with continuous coordinates as inputs and three color channels as outputs. 
However, this approach does not exploit the spatial and temporal redundancy of video signals, and our goal is to improve the parameter efficiency by explicitly removing the redundancy.

We propose \textit{Neural Residual Flow Fields (NRFF)}, a novel neural field scheme for video representation that leverages optical flows and residuals instead of raw colors.
Our proposed scheme was inspired by standard video compression algorithms~\cite{h264,le1991mpeg,hevc} that use motion information to deduplicate signals presented across frames.
Optical flows allow us to reuse color values from other reference frames, which often preserve fine details.
If those delicate patterns are on the surface of subjects inside video frames, reusing color values relieves the burden of learning similar patterns across frames and improves network parameter efficiency, rather than wasting network capacity on storing redundant raw signals for the entire frames.

Many video compression methods use block-wise motion estimations, where each motion vector is applied to all pixels within each block, significantly reducing the total number of required motion vectors.
This is based on prior knowledge that smooth, low-frequency motion vector fields are often sufficient for describing movement between video frames. 
This gave us the idea that substituting optical flows for raw colors may greatly reduce the number of parameters.
However, video frames cannot be completely reconstructed solely by optical flows due to occlusion and dis-occlusion.
Thus, we use \textit{residuals} to recover the original signals from video frames with precision.
It would not necessitate a large neural network since a substantial number of color values are likely to be reused from the motion information.
To summarize, we train the networks to capture optical flows and residuals rather than raw color signals, which greatly improves the efficiency of network parameters.

We also propose to use more than one reference frame for video frame generation.
Using multiple reference frames enables the exploitation of visible information over many reference frames, each of which may contain distinct exposed and occluded information.

In addition, we also suggest splitting the network into two subnetworks: one for optical flows and the other for residuals.
Separating optical flows and residuals, assuming they have different dynamics, would improve the quality, and the experiment results support this.

Experimental results show that the proposed method significantly outperforms the baseline method, which relies on raw colors.
Given similar sizes, \textit{NRFF} reconstructs video frames more clearly and sharply.
Quantitative results on the MPI Sintel~\cite{sintel} and UVG~\cite{UVG} datasets reveal that our method significantly outperformed its counterpart in terms of standard image reconstruction metrics {(PSNR: 31.2 to 37.4, SSIM: 0.82 to 0.95)}.
Although the proposed method is an initial attempt to improve the parameter efficiency of neural fields for videos, it also performs favorably with H.264~\cite{h264}, a standard video compression algorithm.
With the multi-reference frames method, \textit{NRFF} matches the performance of H.264 using small group of pictures (GOP) on some videos without any model compression techniques such as pruning and entropy coding (Fig.~\ref{fig:multi_reference_quant}).

In summary, our contribution is threefold.

$\bullet$ We show that using optical flows and residuals as output instead of colors can significantly improve video quality.

$\bullet$ We propose to use multiple reference frames for frame reconstruction, and this improves video quality without increasing the network size.

$\bullet$ We demonstrate that parameter efficiency can be improved by using separate neural fields—one for optical flows and the other for residuals—in addition to using a shared network for each group of pictures.

\section{Related Works}
\label{sec:related}
\subsubsection{Neural fields}
Neural fields map spatial and temporal coordinates to certain physical quantities~\cite{neuralfields}.
Since a wide variety of tasks can be represented as fields, this approach has recently gained popularity and been used in several tasks, such as image representation~\cite{tancik2020fourier,martel2021acorn,hertz2021sape}, audio representation~\cite{sitzmann2020implicit}, 3D shape~\cite{mescheder2019occupancy,Atzmon_2020_CVPR}, and novel view synthesis~\cite{mildenhall2020nerf,martin2021nerf,zhang2020nerf++,NEURIPS2020_1a77befc,Yu_2021_CVPR}.
Thanks to recent innovations~\cite{tancik2020fourier,sitzmann2020implicit,mildenhall2020nerf,hertz2021sape,barron2021mipnerf}, it can faithfully reconstruct even high frequency signals.
Neural fields have also been applied to signals having both spatial and temporal dimensions, including video representation and novel view synthesis in 4D space~\cite{li2020neural,Gao-freeviewvideo,hao2021nerv,Peng_2021_CVPR,VideoINR}.
As a new way of representing data, several attempts have been made to compress various signals, such as images~\cite{dupont2021coin,strumpler2021implicit,dupont2022coin++} and videos~\cite{hao2021nerv,zhang2021implicit}.
However, the compression performance of neural field-based methods is currently far behind standard state-of-the-art compression algorithms.

\subsubsection{Learning based video compression}
There have been several data-driven attempts to utilize neural networks for efficient video representation.
Convolutional neural networks (CNNs) and auto-encoder architectures have been used to compress video signals~\cite{learned_video_compression,Lu_2019_CVPR_DVC}.
These works train encoder and decoder networks on large-scale datasets and test them on unseen videos to achieve high video compression rates, assuming decoder networks are already shared and only core video information needs to be stored or sent.
DVC~\cite{Lu_2019_CVPR_DVC} has achieved compression rates comparable to or slightly better than standard video compression algorithms, e.g., H.264 and H.265.
However, these approaches are inherently vulnerable to the biases of training datasets.
SIREN~\cite{sitzmann2020implicit} is an attempt to represent various forms of signals, including videos, through neural fields, however, it does not consider its parameter efficiency.
NeRV~\cite{hao2021nerv} is a variant of neural fields that achieves video compression performance comparable to that of the standard video compression algorithm, H.264.
However, NeRV gives up two degrees of freedom and uses only time coordinates $t$ as inputs for efficient rendering.
Our proposed method, \textit{NRFF}, explicitly removes the redundancy by combining residuals and flows with the help of key frames.
IPF~\cite{zhang2021implicit} is a concurrent work using optical flows and residuals for video compression.
Ours and IPF are different in a number of ways.
While IPF employs a separate network for each frame, we use a shared network for each group of pictures and take advantage of temporal redundancy between frames.
We also propose to use more than one reference frame, whereas IPF only proposes to use one.

\subsubsection{Optical flow estimation}
Optical flow has been a core component of various computer vision tasks.
Since the work of Horn and Schunck~\cite{Horn81determiningoptical}, many improvements have been proposed to make optical flow more accurate~\cite{Horn81determiningoptical}.
Recently, employing neural networks to improve optical flow estimation~\cite{flownet,IMKDB17_FLOWNET2.0,Hur:2019:IRR} rather than traditional algorithm-based methods~\cite{Horn81determiningoptical} has been successful.
To achieve more robust learning based optical flow methods, the ground truth optical flows have been collected by using an animated film and computer graphics~\cite{sintel,flownet}.
Several strategies have been proposed to improve estimating performance by using occlusion masks~\cite{zhao2020maskflownet} or transformer-based operations~\cite{teed2020raft,gma_flow}.

\section{Method}

Figure~\ref{fig:main} illustrates an overview of optical flow-based neural fields for video representation. 
Our proposed neural fields generate optical flows for a given spatial and temporal coordinates (Sec~\ref{sec:flow_estimation}) to warp reference frames.
In addition to optical flows, the residuals are also generated by neural fields, and these generated residuals are added to the warped frame to complete the reconstruction (Sec~\ref{sec:image_completion}).
To improve video quality, we use more than one reference frame for video frame reconstruction.
For each GOP, we use two neural networks: one for optical flows and the other for residuals.

\subsection{Dense Optical Flow Estimation}
\label{sec:flow_estimation}
Standard video compression methods use block-wise motion information to improve compression efficiency~\cite{h264,hevc}.
However, because of this algorithmic nature, results often contain block-shaped distortions or artifacts, which necessitate deblocking filters.
We propose using dense, pixel-wise motion vectors through neural fields.
It is known that neural networks can efficiently represent smooth signals~\cite{tancik2020fourier,rahaman2019spectral}.
Thus, we sample motion vectors densely by injecting dense coordinates and using small-size neural networks as neural fields.

\subsection{Image Warping and Completion}
\label{sec:image_completion}
We use generated flow fields to warp reference frames to predict target video frames.
Let $I$ be an image function that takes spatial and temporal coordinates $(x, y, t)$ as inputs and produces corresponding colors $(r, g, b)$ as outputs.
A reference frame and a warped image are denoted as $\hat{I}_{ref}$ and $\tilde{I}$, respectively.
The reference frame can be the key frame or a neighboring frame.
Then, a warped image at time $t$, $\tilde{I}(x,y,t)$ can be written as
\begin{align}
    &(\Delta x^t, \Delta y^t) = F_{\text{flow}}(x,y,t; \theta), \\
    &\tilde{I}(x,y,t) = \text{Interp}(\hat{I}_{ref}(t), x+\Delta x^t, y+\Delta y^t).
\end{align}
$F_{\text{flow}}(t)$, which is parameterized by $\theta$, estimates optical flows between two frames (the video frame to be reconstructed and the reference frame).
$\text{Interp}(\cdot)$ warps a video frame by a bicubic interpolation.

Reconstructing a video frame by simply warping the source video frame is likely to contain artifacts.
Artifacts can be caused by a variety of factors, including imprecise optical flow estimation, occlusions and disocclusions between video frames, and accumulated errors from the recursive frame generation process.
To alleviate the issues, we use residuals along with the optical flows.

The final equation for image completion can be written as
\begin{equation}
\label{eq:image_competion}
\hat{I}(x, y, t) = F_{\text{res}}(x,y,t; \psi) + \tilde{I}(x, y, t)
\end{equation}
, where $F_{\text{res}}$ denotes the residual estimator with its own parameter $\psi$.

\subsection{Key Frames}
\label{ssec:keyframe}

Inspired by the standard video compression algorithms, we store a key frame per GOP as a standalone frame so that other frames directly or indirectly depend on the key frame.
Regarding key frames, there are two important considerations: the image quality of key frames and the location of key frames.

With a fixed total size, there is a trade-off between the key frame quality and the network size.
Given a high-quality key frame, a network can exploit the high-frequency details of the key frame.
However, since high-quality key frames require a larger memory size, the network size must be smaller in order to maintain the total size.
Having a small-sized network tends to have difficulties covering long, dynamic frames.
In contrast, a relatively low-quality key frame relegates expressing fine-details to the network.
A large network handles long dynamic frames relatively easily, however, because the reference frames lack fine-details, the network must learn to compromise between learning fine-details and learning the optical flows of long and dynamic frames.
We empirically found that the optimal ratio of the network size and the key frame quality (or the size) is related to the total number of frames in the GOP.
The larger the GOP is, the larger the network size should be, and vice versa. 
The detailed experimental results can be found in the supplementary material.

Among possible positions, we chose the middle frame as the key frame to minimize the errors caused by the distance from the key frame.
Experimental results showed that selecting the middle frame as the key frame is better in terms of reconstruction quality than selecting the first or last frame in the GOP.

We could use neural fields for key frame compression.
However, existing neural field-based image compression methods usually underperform or are at most similar to standard image compression algorithms, such as JPEG, in terms of compression efficiency.
Therefore, for keyframe compression, we employ the standard H.264 video compression technique.
We still need more sophisticated training algorithms, input feature preprocessing techniques, and new network architectures to completely replace H.264 with neural fields in our proposed method.

\subsection{End-to-end Training}

Our proposed approach was designed to be differentiable throughout the entire process in order to build an end-to-end framework for video representation learning.
We used MSE (mean squared error) loss to minimize the reconstruction errors.
Let $B$ be a mini-batch of frame indices excluding the key frame, then the loss function can be written as
\begin{equation}
\label{eq:total_loss}
L(\theta, \phi, \psi) = \frac{1}{Z}\sum_{t \in B}\sum_{x}\sum_{y} ||I(x, y, t) - \hat{I}(x, y, t)||^2.
\end{equation}
$I(\cdot,\cdot,\cdot)$ is a ground truth color, and $\hat{I}(\cdot,\cdot,\cdot)$ is a reconstructed color.
We multiplied the equation with the inverse of the constant $Z$ to get the average loss over both time and space.
Note that we excluded the key frame during the training process.
The key frame will only be included during the evaluation.

\begin{figure*}[t]
\centering
\includegraphics[width=1\textwidth]{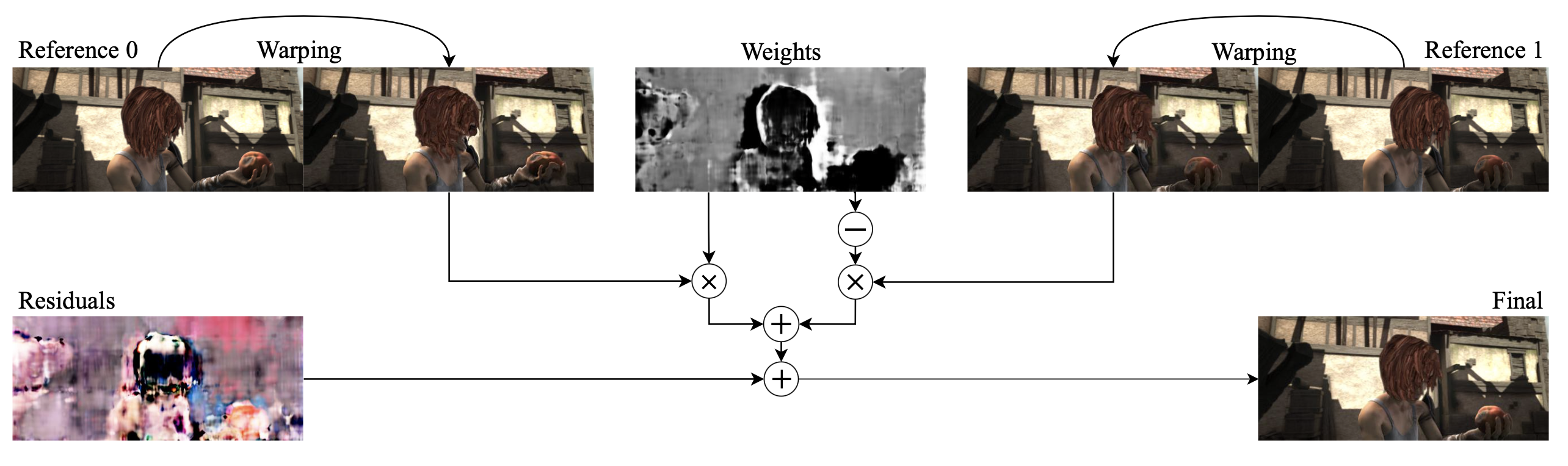}
\caption{Neural residual flow fields with multi-reference frames. $\otimes$ and $\oplus$ denote element-wise multiplication and addition, respectively. $\ominus$ in the figure denotes one minus inputs, that is, in the case of weights, one minus weights will be outputted.}
\label{fig:multi_main}
\end{figure*}

\subsection{Multi-reference Frames}
To further improve the video quality, we propose to use more than one reference frame to reconstruct a video frame.
For each video frame, the video representation network learns to warp the nearest two key frames, each of which is unique per GOP, to reconstruct a video frame.
The key frame from the GOP, where a particular video frame to be reconstructed is located, is always referred to for frame reconstruction.
A frame, preceding the key frame of the GOP in temporal order, uses the key frame from the nearest preceding GOP as the other reference.
Likewise, a frame after the key frame will use the key frame of the nearest posterior GOP as another reference frame.

Overall architecture is presented in Fig.~\ref{fig:multi_main}.
First, two reference frames are warped by the generated optical flows.
To combine information from multiple frames, we also need learned pixel-wise weights (from zero to one) that selectively aggregate two warped frames.
The network would weigh more on closer pixels among two reference frames to aggregate warped images.
The flow network now generates optical flow outputs for each reference frame and additional mixing weights.
Lastly, the residuals are added to fully reconstruct a video frame.

\subsection{Network Split}
\label{ssec:split}

Simply replacing the output of neural fields from colors with optical flows and residuals means that optical flows and residuals are generated from the shared parameters.
If the patterns of optical flows and residuals are similar, then sharing parameters may help to improve parameter efficiency.
Otherwise, sharing parameters could degrade the quality.
We analyze two different structures in the experimental section.

As proposed in IPF, we can also split in the temporal dimension.
That is, we can use each network for each video frame.
However, unlike the relationship between optical flows and residuals, neighboring video frames are usually highly correlated.
To support our claim that separating the network for each video frame prevents the use of temporal redundancy and, thus, is less parameter efficient, we compare our method with IPF in the experimental section.

\section{Experiments}
We tested our approach on both synthetic and real-world datasets (MPI Sintel~\cite{sintel}, UVG dataset~\cite{UVG}).
We used the color-based neural field, SIREN~\cite{sitzmann2020implicit}, as the baseline.
In addition to SIREN, we also compared our approach with the standard video compression algorithm, H.264~\cite{h264}.

\begin{figure}[t]
\centering
\includegraphics[width=1\textwidth]{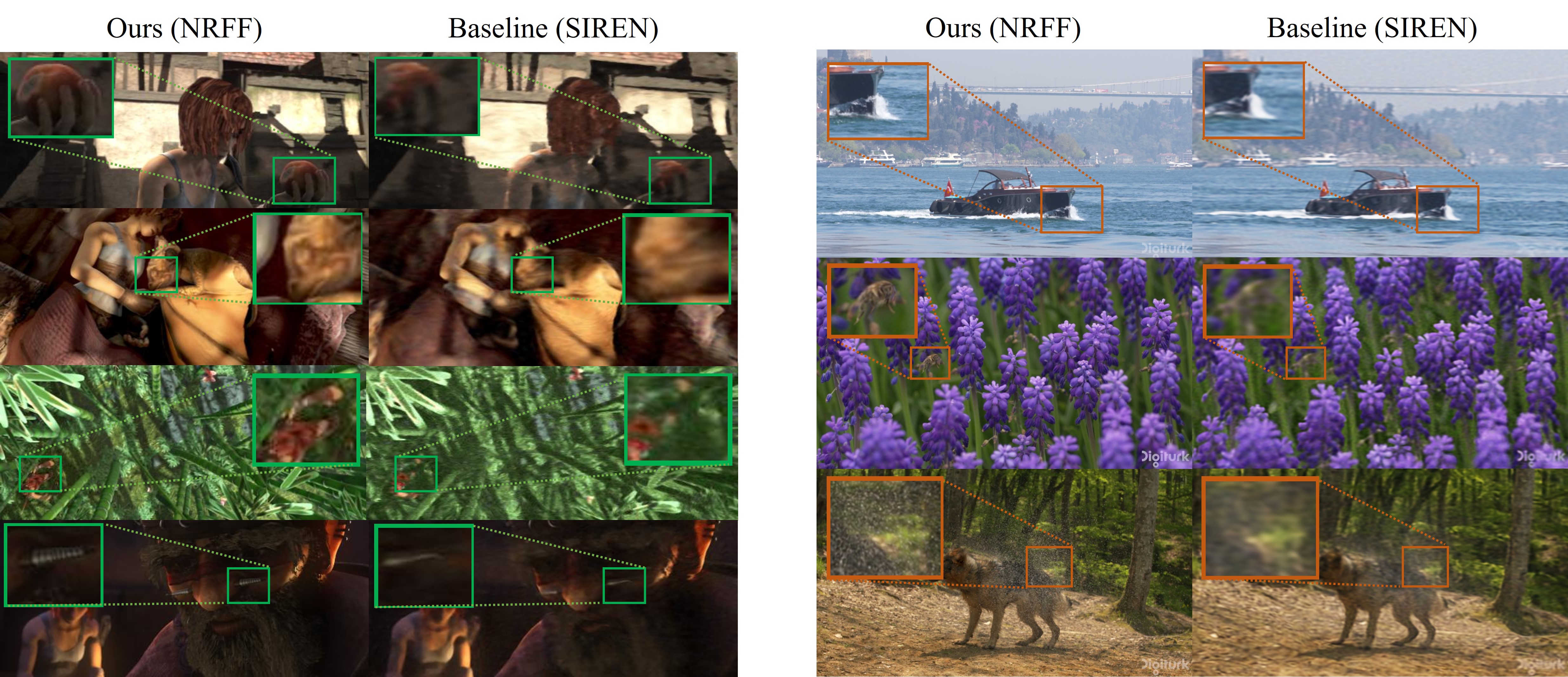}
\caption{The qualitative results between the proposed method and the baseline method under similar sizes (best viewed in color electronically). The left column examples are from the MPI Sintel dataset, and the right column examples are from the UVG dataset.}
\label{fig:baseline_qualitative}
\end{figure}

\begin{figure*}[t]
\centering
\includegraphics[width=\textwidth]{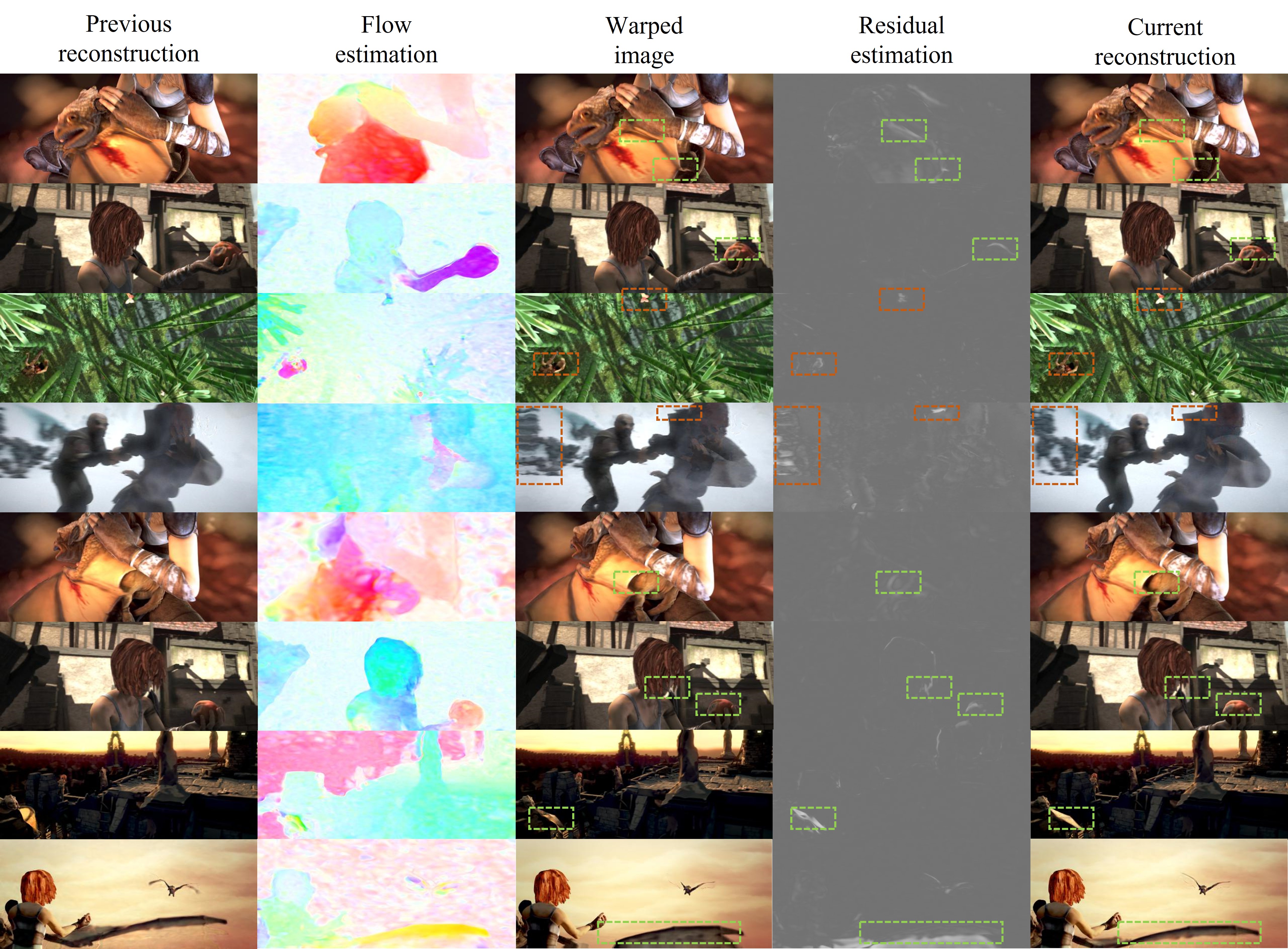}
\caption{The qualitative results on the MPI Sintel dataset.}
\label{fig:mpi_qualitative}
\end{figure*}

\subsection{Dataset}

The MPI Sintel~\cite{sintel} dataset was originally designed to evaluate the optical flow performance and it provides challenging and natural video sequences based on an open source animated film.
We selected the MPI Sintel dataset for two reasons.
First of all, it provides the ground truth optical flow, which allows us to evaluate the accuracy of the learned flows.
Flow estimation is a core part of our algorithm, therefore, it is desirable to understand how flow estimation affects the final reconstruction.
Second, it has been a good testbed for challenging scenarios, such as long range motion, motion blur, multi-frame analysis, and non-rigid motion.
We tested our method on the entire 23 videos supported by MPI Sintel.

We also evaluated our method using seven 1080p videos from the Ultra Video Group (UVG) video collection~\cite{UVG}, in order to compare it to other neural field-based video compression methods~\cite{hao2021nerv,zhang2021implicit}.
The dataset includes a variety of videos, ranging from nearly static scenes to fast-moving scenes.

\subsection{Single Reference Frame Experiments}
\subsubsection{Experimental setup}
First, we used the SIREN backbone model and only changed the output linear layer to evaluate the effect of replacing colors with optical flows and residuals.
For frame reconstruction, we applied optical flows recursively; that is, starting from the key frame, the reconstructed frame is used as the reference frame of the next frame.
We used 16-bit precision weights for both ours and the baseline (SIREN) to halve the network size without quality degradation.

Our proposed method and the baseline are evaluated using the MPI Sintel video dataset.
The resolution of all videos was reduced by half, resulting in $218 \times 512$ pixels.
We set the size of the GOP to seven, so that each key frame covers six neighboring frames.
In the case of a video containing 28 frames, for example, we trained four models for the video.
We trained each model for total 30K iterations, and training the whole batch of frames at once was considered a single iteration.
We used the Adam optimizer with a learning rate of 0.0005.


\begin{figure*}[t]
\centering
\includegraphics[width=1\textwidth]{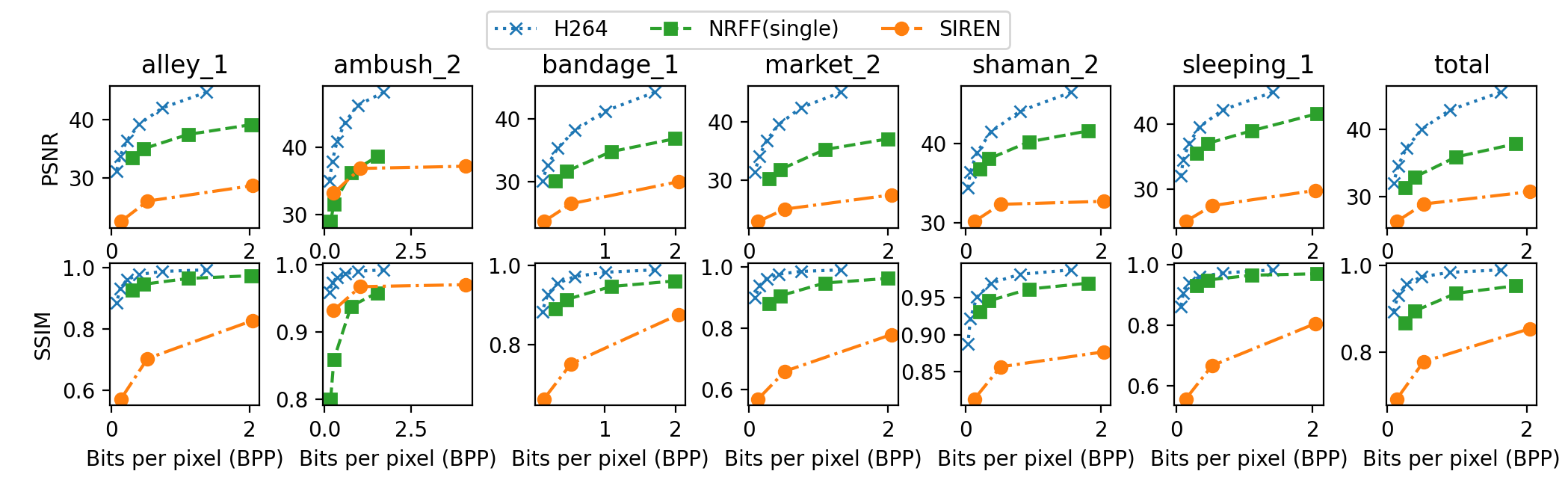}
\caption{The quantitative results of single reference frame experiments on the MPI Sintel dataset.}
\label{fig:mpi_quant}
\end{figure*}

\subsubsection{Results}
As shown in Fig.~\ref{fig:baseline_qualitative}, replacing colors with optical flows and residuals significantly improves video quality for similar model sizes.
Our method effectively preserves fine details such as a person's beard, water drops, and bees, whereas the baseline method generates blurry outputs.

Figure~\ref{fig:mpi_qualitative} shows the qualitative results of single reference frame experiments in detail, including intermediate steps to complete video frames.
The parts that need to be corrected by residuals are highlighted in figure~\ref{fig:mpi_qualitative}.
As we expected, the optical flow estimation is not necessarily accurate enough to reconstruct a video frame correctly.
In fact, the final optical flow estimation includes many artifacts, particularly in the background regions.
The residuals can successfully compensate for those artifacts created by inaccurate optical flows.

Due to the page limits, we presented the results of six videos and the average of all 23 videos in Fig.~\ref{fig:mpi_quant}.
We used two commonly used metrics in image reconstruction tasks; peak signal noise ratio (PSNR) and structural similarity index measure (SSIM).
On average, our method enhanced PSNR from 30 to 37 and SSIM from 0.85 to 0.95 at around 2 bits per pixel. 
For example, in \textit{alley\_1}, we obtained 39.09 PSNR and 0.972 SSIM, as opposed to the baseline performance of 28.71 PSNR and 0.827 SSIM under similar model sizes.
In \textit{sleeping\_1}, we got 41.56 PSNR and 0.970 SSIM, while the baseline only reached 29.78 PSNR and 0.806 SSIM.
As illustrated in Figure~\ref{fig:baseline_qualitative}, this resulted in significant increases in visual quality.

There are a few exceptions where we did not gain much improvement.
One example is demonstrated in the fourth row of Figure~\ref{fig:mpi_qualitative}.
The scene is foggy and blurry, and the video shows a lot of camera movement.
These factors, we believe, are the causes of the slightly lower quality of our approach.
We hypothesize that, given a fixed number of parameters, inaccurate optical flow estimation may not be sufficiently compensated by residuals in some scenes.
We also reported the performance of the H.264 video compression method~\cite{h264} for your information.

\begin{figure}[t]
\centering
\includegraphics[width=1\textwidth]{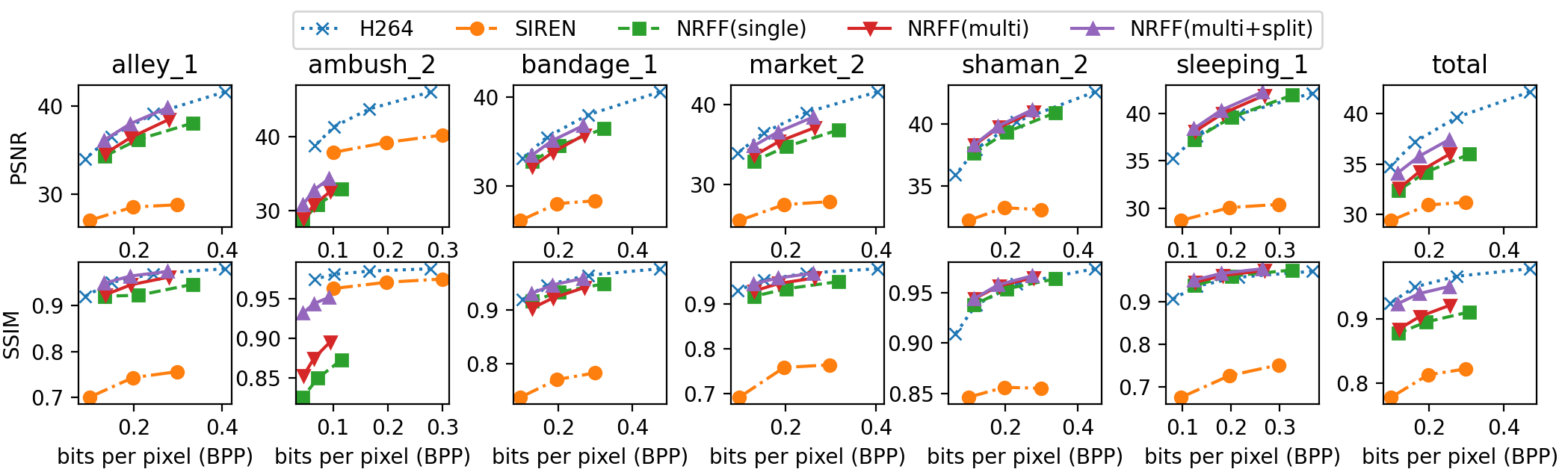}
\caption{The quantitative results of multi-reference frames experiments on the MPI Sintel dataset. The last column is the average performance on all 23 videos.}
\label{fig:multi_reference_quant}
\end{figure}

\subsection{Multi-reference Frames and Network Splitting Experiments}
\subsubsection{Experimental setup}
This section analyzes the effect of using multi-reference frames and separating networks on video quality.
We replaced the activation function of the neural networks from sinusoidal activation~\cite{sitzmann2020implicit} to swish function ($f(x) = x \times \mathsf{sigmoid}(\beta x)$) and used positional encoding~\cite{mildenhall2020nerf}.
Furthermore, we reused hidden layers to improve parameter efficiency.
We used H.264 for key frame compression, and the middle frame of the GOP was chosen as the key frame.
To modulate the total size for video representation, we controlled the quality factor in H.264.
As for the network size, each network width was automatically set so that the total size of the network is proportional to the corresponding key frame size.

We ran experiments using full-resolution MPI Sintel videos.
We set the size of the GOP to five for every method, including H.264.
We also applied 16 bit precision in this experiment to both the baseline (SIREN) and our proposed methods for a fair comparison.
We trained each model for 5K epochs, and the batch size was set to one.
Since the single reference NRFF becomes unstable when the batch size is one, the whole batch of the GOP was used as the minibatch for the single reference NRFF and the number of epochs was therefore increased to 25K.
As for the initial learning rate, we used optimal learning rate for each method: 1e-3 for SIREN and the single reference NRFF, and 1e-2 for the multiple reference NRFF.
The NRFF network size was set to be equal to one-fourth of the key frame size.
For example, a network with a key frame size of 10k bytes was set to have approximately 1,250 parameters of 16-bit precision.

We also evaluated each method using videos from the UVG dataset (the first 100 frames of each).
In this experiment, we unbridled the GOP size limitation for SIREN and H.264.
That is, we use the automatically chosen GOP size in the case of H.264, which is 100, and we use the same size for SIREN.
For our method, the GOP size was set at 15.
Since we empirically found that large GOP size requires an increased ratio of the network size and the key frame, we set the ratio to be approximately 0.5.

\subsubsection{Results}
\begin{figure}[t]
\centering
\includegraphics[width=1\textwidth]{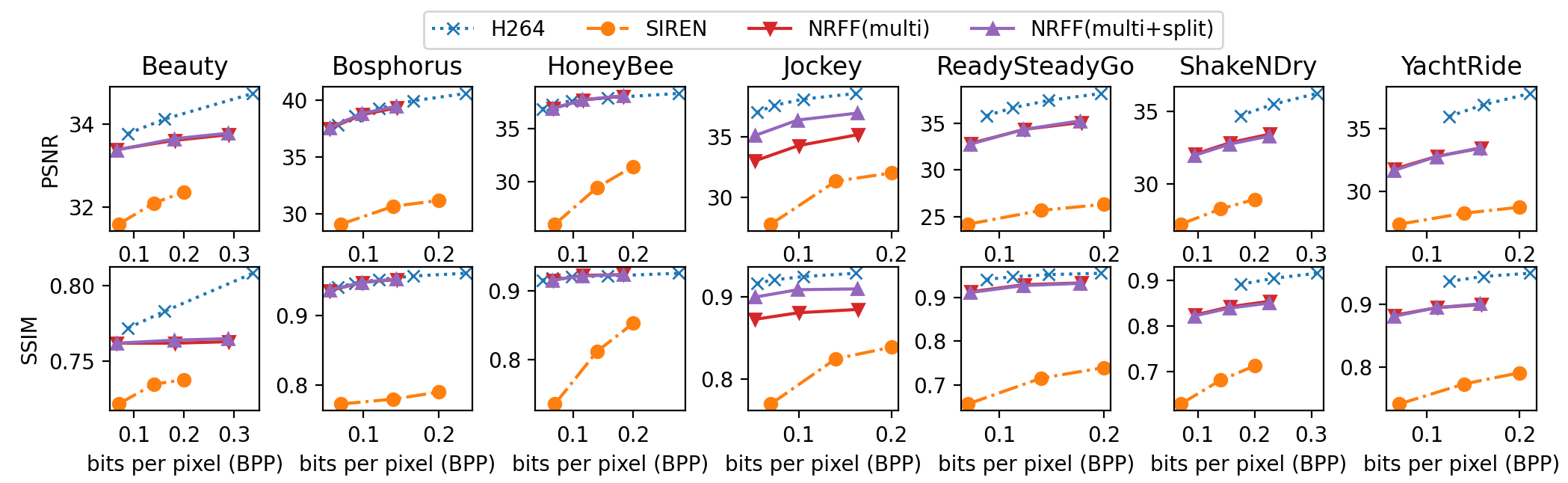}
\caption{The quantitative results of multi-reference frames experiments on the UVG dataset.}
\label{fig:multi_reference_uvg_quant}
\end{figure}

Figs.~\ref{fig:multi_reference_quant} and \ref{fig:multi_reference_uvg_quant} show the quantitative results on the MPI Sintel and UVG datasets, respectively.
The supplementary materials provide all of the experiment results on the MPI Sintel.
First of all, the results indicate that using optical flows can enhance video quality for both synthetic and non-synthetic scenes, including static and dynamic scenes.
Second, referencing more than one frame enhances video quality without increasing the network size.
Lastly, dividing the network into two subnetworks never degrades video quality and can significantly improve quality for some videos.
This demonstrates that our assumption that optical flows and residuals have different dynamics is valid.
Even without network compression methods, our method performs on par with or better than the H.264 algorithm on some videos, such as \textit{shaman\_2} and \textit{sleeping\_1} from MPI Sintel, and Bosphrous and HoneyBee from UVG.
The optical flow-based method might worsen the video quality in some videos, such as \textit{ambush\_2}, which contains fast and large movements.
However, the overall performance can be significantly improved by using optical flows instead of raw colors, as shown in Fig.~\ref{fig:multi_reference_uvg_quant}, and the rightmost column of Fig.~\ref{fig:multi_reference_quant}



\begin{figure}[t]
\centering
\begin{subfigure}[b]{0.49\linewidth}
    \centering
    \includegraphics[width=\linewidth]{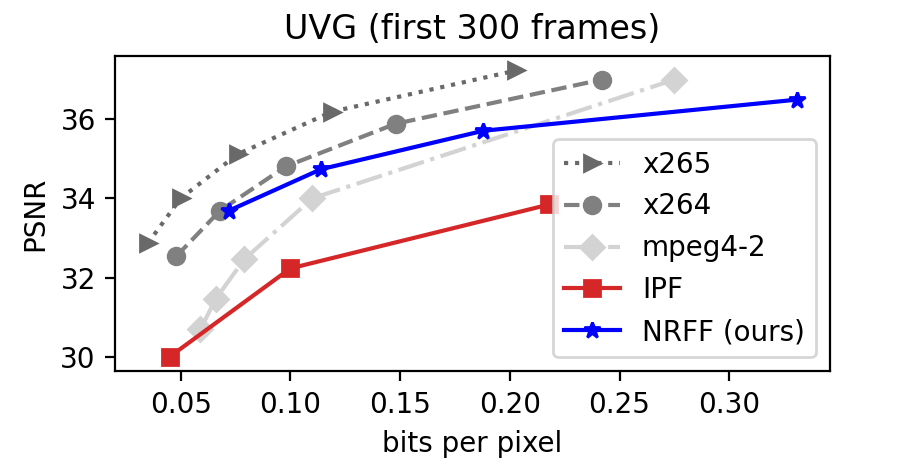}
    \caption{comparison with IPF}
    \label{fig:sub_zhang}
\end{subfigure}
\begin{subfigure}[b]{0.49\linewidth}
    \centering
    \includegraphics[width=\linewidth]{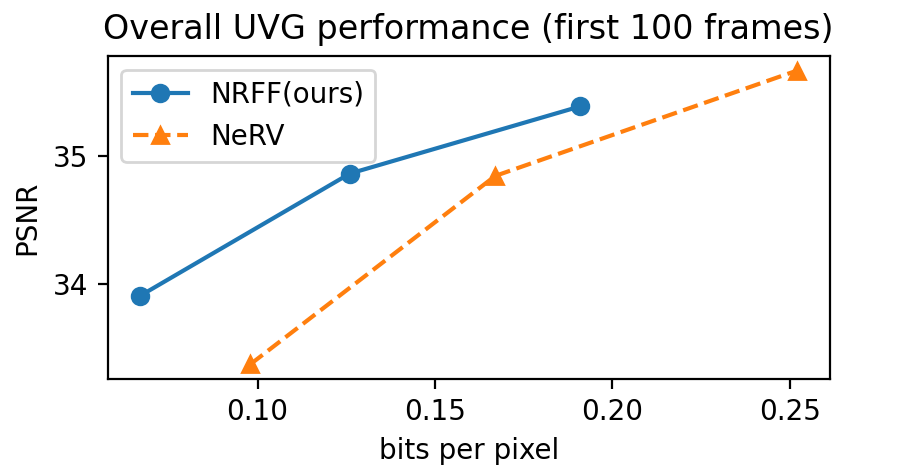}
    \caption{comparison with NeRV}
    \label{fig:sub_nerv}
\end{subfigure}
\caption{Overall compression performance on the UVG videos. All results in fig.~\ref{fig:sub_zhang} were copied from IPF~\cite{zhang2021implicit} except NRFF (ours).}
\label{fig:nrff_nerv}
\end{figure}

\subsection{Comparison with Other Neural-field-based Video Representation}
\subsubsection{Experimental setup}
We compared our method (multi-reference model with two subnetworks) to other neural field-based methods (IPF~\cite{zhang2021implicit}, NeRV) using the UVG dataset.
Since these two methods are evaluated in different experimental settings, we separately compared them with ours.
First, to compare ours with IPF, we used the same experimental settings as in IPF (first 300 frames with the size of the group of pictures (GOP) of five) to train our models.
To compare ours with NeRV, we compared the results of publicly available NeRV codes in our experimental settings (first 100 frames for 1,500 epochs, and no compression techniques other than 16-bit precision).

\subsubsection{Results}

Fig.~\ref{fig:nrff_nerv} shows how efficiently our proposed neural fields can express a video compared to other neural field-based methods, including the current state-of-the-art method, NeRV.
Our method outperforms a concurrent flow-based neural fields, IPF by a large margin, as demonstrated in fig.~\ref{fig:sub_zhang}.
This performance gap is significant considering that IPF has adopted additional compression techniques, such as quantization and entropy encoding, to improve compression performance, while ours does not.
We conjecture that this performance gain pertains to using a shared network per GOP.

Fig.~\ref{fig:sub_nerv} shows that replacing raw colors with residuals and flows results in better performance, especially in the low bits per pixel (bpp) region, even without enhancing network structure.
Please note that NeRV relies on a neural field architecture that limits spatial sampling and only allows temporal sampling to improve compression performance, while ours permits spatial sampling.

\subsection{Spatial and Temporal Interpolation}
The neural fields have several advantages that no other video compression method has, one of which being the ability to sample values from arbitrary spatial and temporal coordinates, even at an unobserved point during encoding.
To show this advantage of neural fields, we ran two experiments: spatial and temporal interpolation.
For spatial interpolation, we first trained a neural network to represent a low-resolution video (with a resolution of (480, 270)).
After that, without any post-processing methods, we simply upscaled the resolution four times in both height and width by sampling values in a much more dense grid.
This is possible because neural fields take continuous coordinates as inputs.
For temporal interpolation, we trained a neural network as in the main experiment with the fixed reference frames as for the multi-reference frame experiments.
And then, the intermediate frame (for example, a frame in between the 5th and 6th frame) was sampled simply by injecting the corresponding temporal coordinates.

\begin{figure} 
\centering
\includegraphics[width=0.7\linewidth]{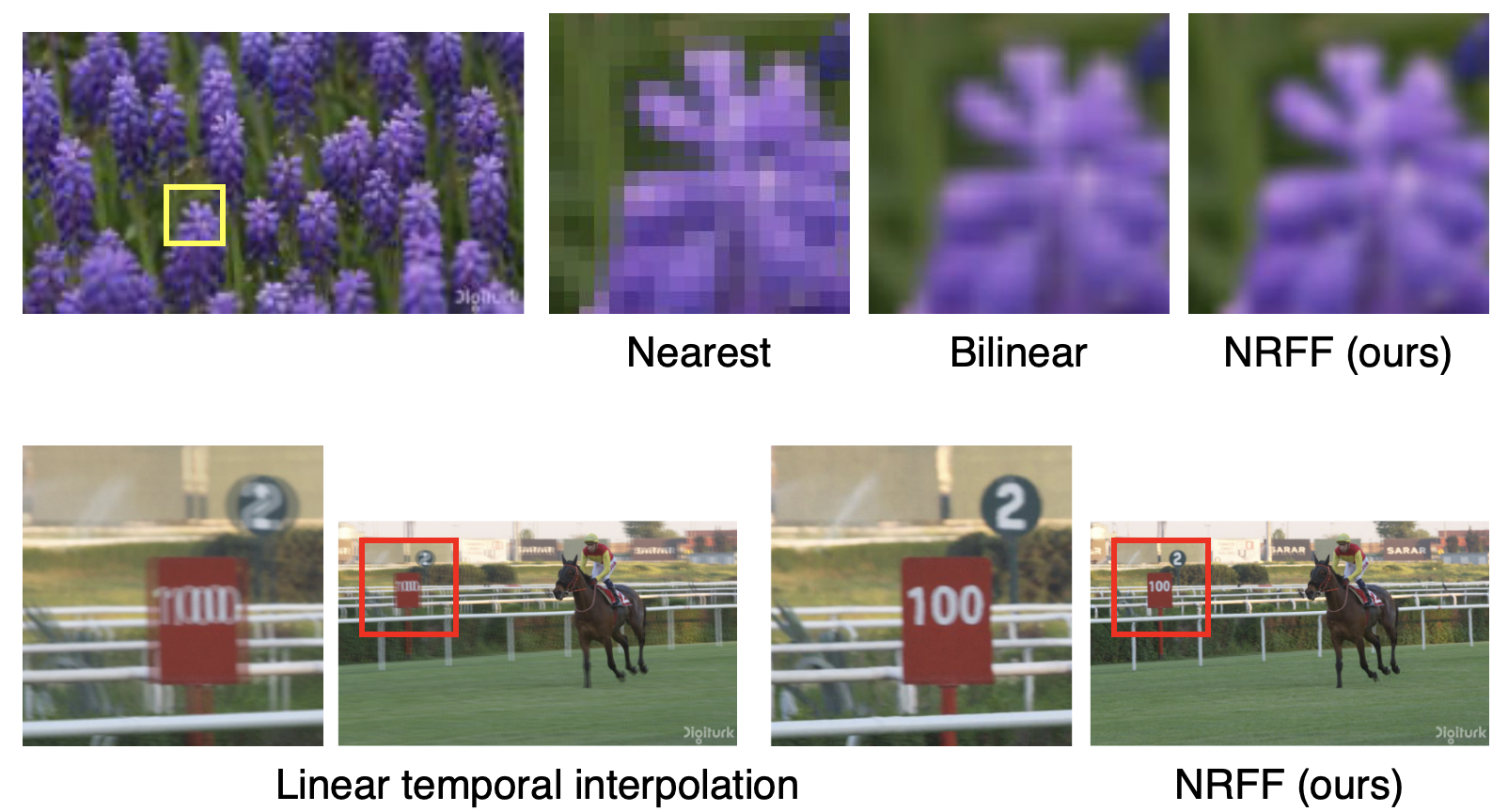}
\caption{The spatial and temporal interpolation.
The first row shows the results of each method's spatial super resolution, upscaling from (480, 270) to (1920, 1080).
The second row shows the the results of temporal interpolation, which samples an intermediate frame that was not seen during video encoding.}
\label{fig:upsampling}
\end{figure}

As shown in Fig.~\ref{fig:upsampling}, simple dense sampling results in smooth interpolation in both time and space, even without any modifications and extra training techniques.
Spatial interpolation of neural fields results in much smoother outputs than bilinear interpolation.
For a fast-moving scene, interpolating two adjacent frames results in a blurry frame.
However, our proposed method manages to represent the intermediate frame much more clearly.
In addition to the fact that NRFF inherits the good properties of neural fields, the proposed method can offer new opportunities to improve the parameter efficiency of neural fields in other domains, such as NeRF~\cite{mildenhall2020nerf} and light-field imaging, to name a few.

\section{Conclusions and Discussion}
\label{sec:conclusion}

We present a way to exploit neural fields for efficient video representations.
The video quality was greatly enhanced by explicitly leveraging reference frames through optical flows. 
The proposed approach, \textit{Neural Residual Flow Fields (NRFF)} maintains smooth and less complex signals, which allows us to achieve more compact representations while maintaining quality.

Although the results were promising, there is still room for improvement.
We observed that neural fields (or implicit neural representations) require a large number of parameters and long training iterations to capture high-frequency details.
We believe that resolving this problem would significantly enhance encoding speed and make neural fields more accessible in various cases.

Note that we achieved fairly good compression rates without any model compression techniques, except for 16-bit precision weights.
Incorporating unaccommodated network compression techniques into our proposed method would improve the performance much further.
Weight pruning, entropy coding, and knowledge distillation could all be promising directions to investigate.

There are a handful of promising research directions that can further make this emerging representation more attractive.
We believe we have only scratched the surface.
A better understanding of the efficiency of neural network parameters in general might help answer a fundamental question about how a neural network preserves information in its parameters.

\subsubsection{Acknowledgements} This research was supported by the Ministry of Science and ICT (MSIT) of Korea, under the National Research Foundation (NRF) grant (2021R1F1A1061259, 2022R1F1A1064184), Institute of Information and Communication Technology Planning Evaluation (IITP) grants for the AI Graduate School program (IITP-
2019-0-00421), the ITRC (Information Technology Research Center) support program (IITP-2021-0-02052), the ICT Creative Consilience program (IITP-2020-0-01821), and the Artificial Intelligence Innovation Hub program (IITP-2021-0-02068).

\clearpage


\renewcommand\thesection{\Alph{section}}
\setcounter{section}{0}

\section{Appendix}

\subsection{Performance Comparison using MPI SINTEL Videos}
\label{sec:sintel}

\subsubsection{Experiment Setup}
\label{ssec:sintel_exp}
In this section, we present all the measured performance on all 23 MPI SINTEL~\cite{sintel} videos of five methods: H.264~\cite{h264}, color-based baseline (SIREN~\cite{sitzmann2020implicit}), our approach with a single reference, multiple references, and lastly, multiple references with separate optical flow and residual models.
We used every frame in each video at its original resolution (436 x 1024).
We set the size of the group of pictures (GOP) to five for all methods.

In this work, we used H.264 for key frame compression in order to compare the compression performance of NRFF with H.264.
Given a key frame (I-frame), H.264 encodes optimal block-wise flows and residuals, and NRFF uses a neural network to compress pixel-wise flows and residuals.
Since key frames are encoded in the same way as in H.264, we can compare how well each method compresses flows and residuals.
This also enables performance comparison between a single reference and multiple reference NRFFs.
When comparing the compression performance of those five methods, keep in mind that we did not apply any network compression methods to neural field-based methods.

\begin{figure*}
    \centering
    \includegraphics[width=0.95\textwidth]{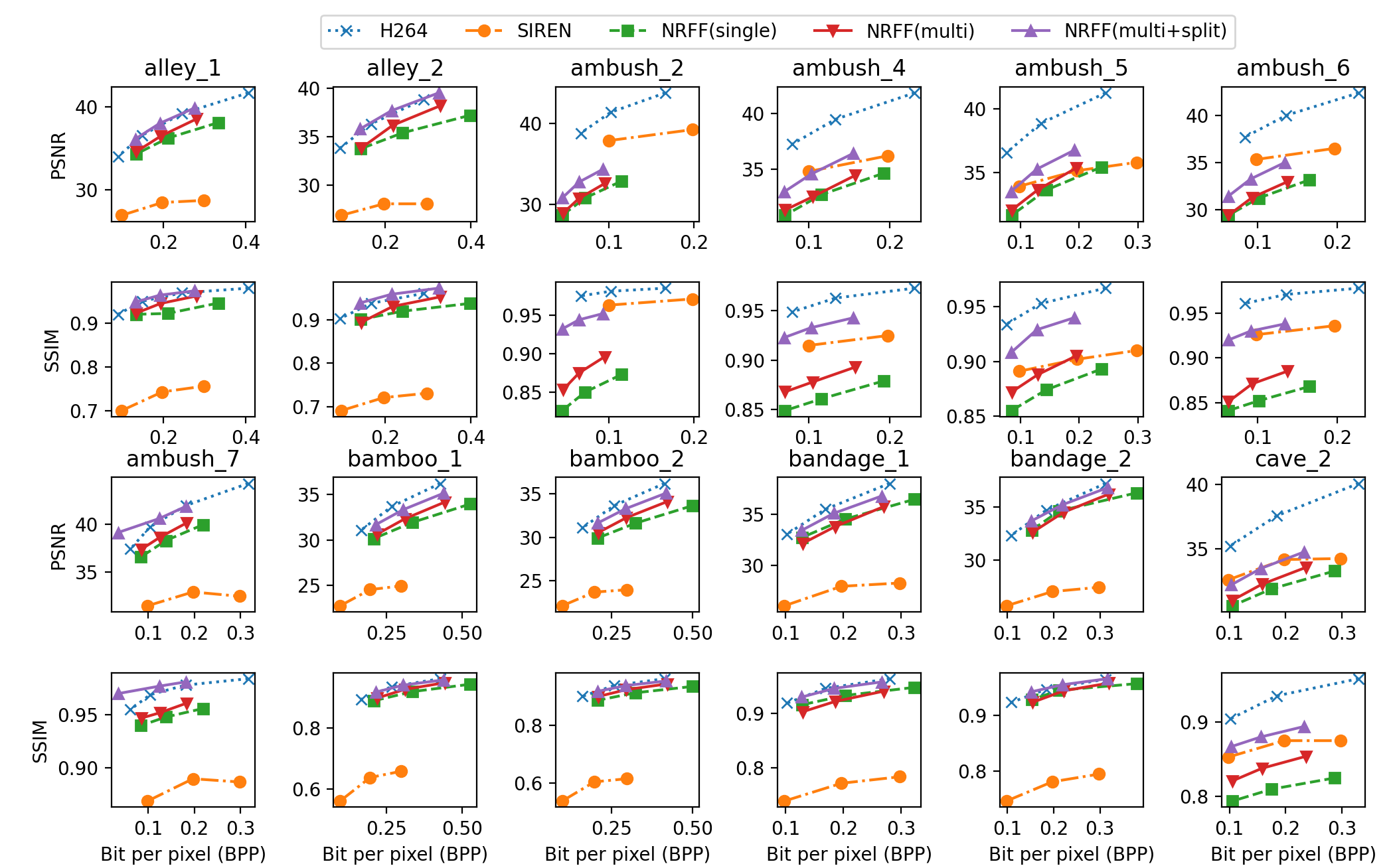}
    \caption{Performance Comparison using MPI SINTEL Videos (1/2)}
    \label{fig:sintel}
\end{figure*}

\begin{figure*}
    \centering
    \includegraphics[width=0.95\textwidth]{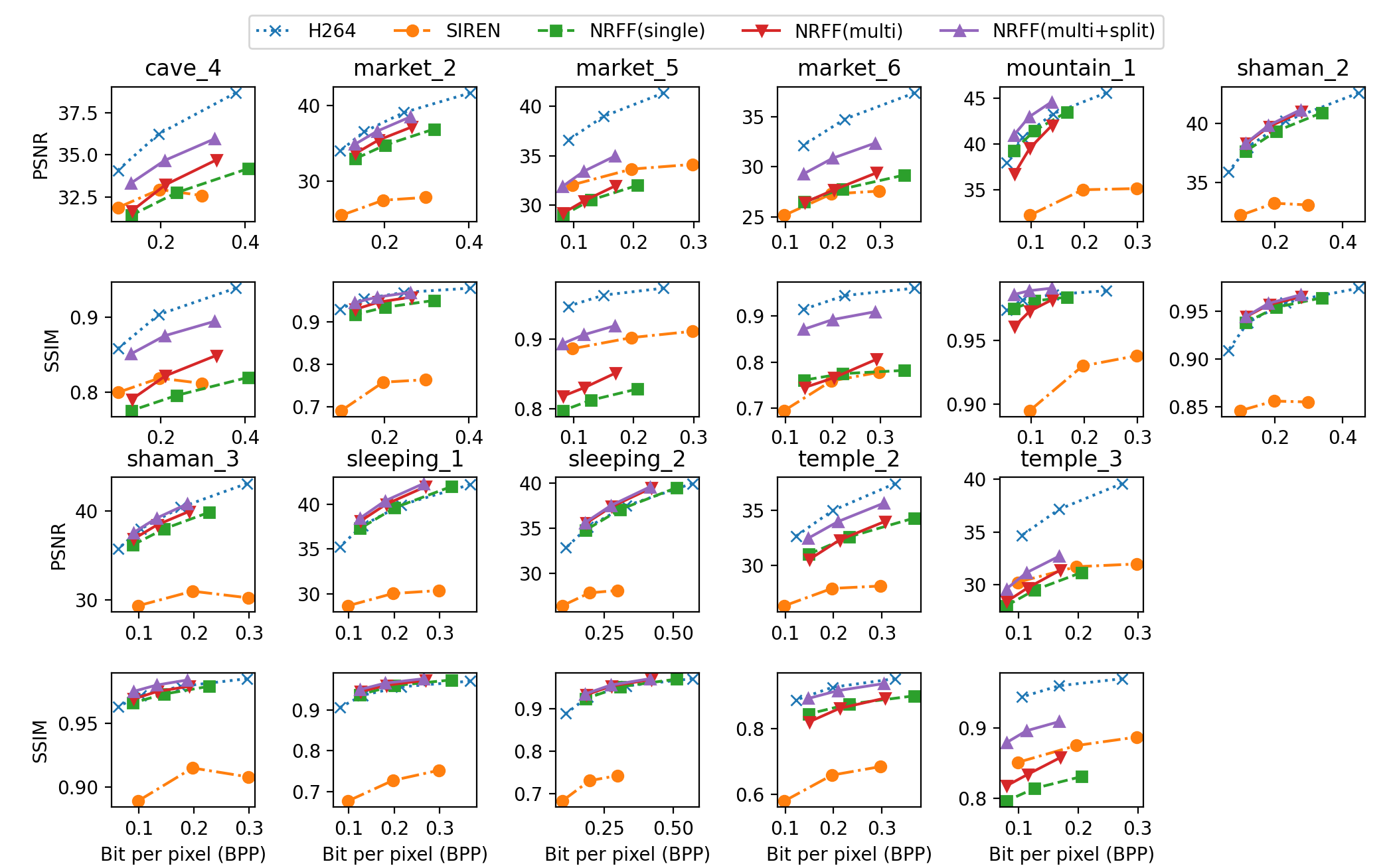}
    \caption{Performance Comparison using MPI SINTEL Videos (2/2)}
    \label{fig:sintel}
\end{figure*}

\subsubsection{Results}
\label{ssec:sintel_results}
Fig.~\ref{fig:sintel} shows the results of five methods on MPI SINTEL videos.
We measured the performance using PSNR and SSIM.
The x-axis of each graph is bits per pixel (bpp).
On most videos, optical flow and the residual-based approach (NRFF) perform better than or at least similar to the color-based approach (SIREN), and on more than half of the videos, it outperforms by a large margin.
NRFF shows lower video quality in \textit{ambush\_2} or \textit{ambush\_6}, which are characterized by the abrupt appearance or rapid movement of an object that spans a substantial amount of visual area.
Regarding the number of reference frames, a single reference NRFF showed inferior performance compared to the multiple reference version even with five times longer training time.
Splitting the network improved video quality in most cases and, surprisingly, never degraded video quality.
The assumption that optical flows and residuals have different dynamics appears to be supported by these results.

\begin{figure}
\centering
\includegraphics[width=0.75\linewidth]{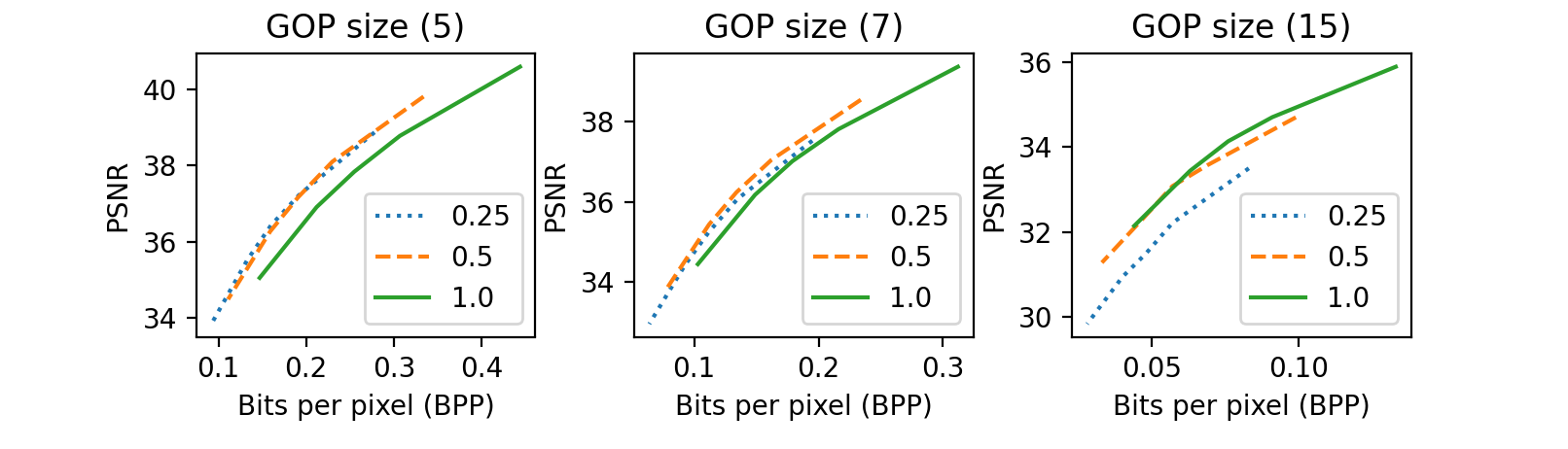}
\caption{The rate distortion curves with different network and GOP sizes in Alley\_1. The size of group of pictures (the total number of frames for each group of pictures) are shown inside the parenthesis. 0.25, 0.5, 1.0 are ratios between network size and the keyframe size.} 
\label{fig:gop}
\end{figure}


\subsection{The ratio of network size and keyframe size}
As shown in Fig.~\ref{fig:gop}, we found that the optimal ratio of network size and keyframe size is proportional to the size of the group of pictures. For example, the ratio of 0.25, which means the network size is one quarter of the key frame size, was optimal in a small GOP, while the larger GOP requires a much higher ratio.

\subsection{Batch Size}
Due to the fact that our proposed method does not restrict the range of optical flows, the training process for some videos may be unstable.
We solved this issue by increasing the batch size to be more than one.

\bibliographystyle{splncs}
\bibliography{egbib}

\end{document}